\documentclass[final]{cvpr}

\usepackage{times}
\usepackage{epsfig}
\usepackage{graphicx}
\usepackage{amsmath}
\usepackage{amssymb}
\usepackage{color}
\usepackage[pagebackref=true,breaklinks=true,colorlinks,bookmarks=false]{hyperref}
\hypersetup{
    colorlinks=true,
    filecolor=magenta,      
    urlcolor=magenta,
}

\def\cvprPaperID{7842} 
\def\confYear{CVPR 2021}
\begin{document}

\title{Multi-Perspective LSTM for Joint Visual Representation Learning}

\author{Alireza Sepas-Moghaddam$^1$ ~~~~~~~~Fernando Pereira$^2$~~~~~~~Paulo Lobato Correia$^2$~~~~~~~Ali Etemad$^1$\\
$^{1}$Dept. ECE and Ingenuity Labs Research Institute, Queen's University, Kingston, Ontario, Canada\\
$^{2}$Instituto de Telecomunicações, Instituto Superior Técnico - Universidade de Lisboa, Lisbon, Portugal\\
{\tt\small \texttt{alireza.sepasmoghaddam@queensu.ca, \{fp, plc\}@lx.it.pt, ali.etemad@queensu.ca}}
%
}



\maketitle

\begin{abstract}

We present a novel LSTM cell architecture capable of 
learning both intra- and inter-perspective relationships available in visual sequences captured from multiple perspectives. Our architecture adopts a novel recurrent joint learning strategy that uses additional gates and memories at the cell level. We demonstrate that by using the proposed cell to create a network, more effective and richer visual representations are learned for recognition tasks. We validate the performance of our proposed architecture in the context of two multi-perspective visual recognition tasks namely lip reading and face recognition. Three relevant datasets are considered and the results are compared against fusion strategies, other existing multi-input LSTM architectures, and alternative recognition solutions. The experiments show the superior performance of our solution over the considered benchmarks, both in terms of recognition accuracy and complexity. We make our code publicly available at \href{https://github.com/arsm/MPLSTM}{https://github.com/arsm/MPLSTM}.
\end{abstract}

%


\section{Introduction}

\begin{figure}[!t]
    \begin{center}
    \includegraphics[width=0.95\linewidth]{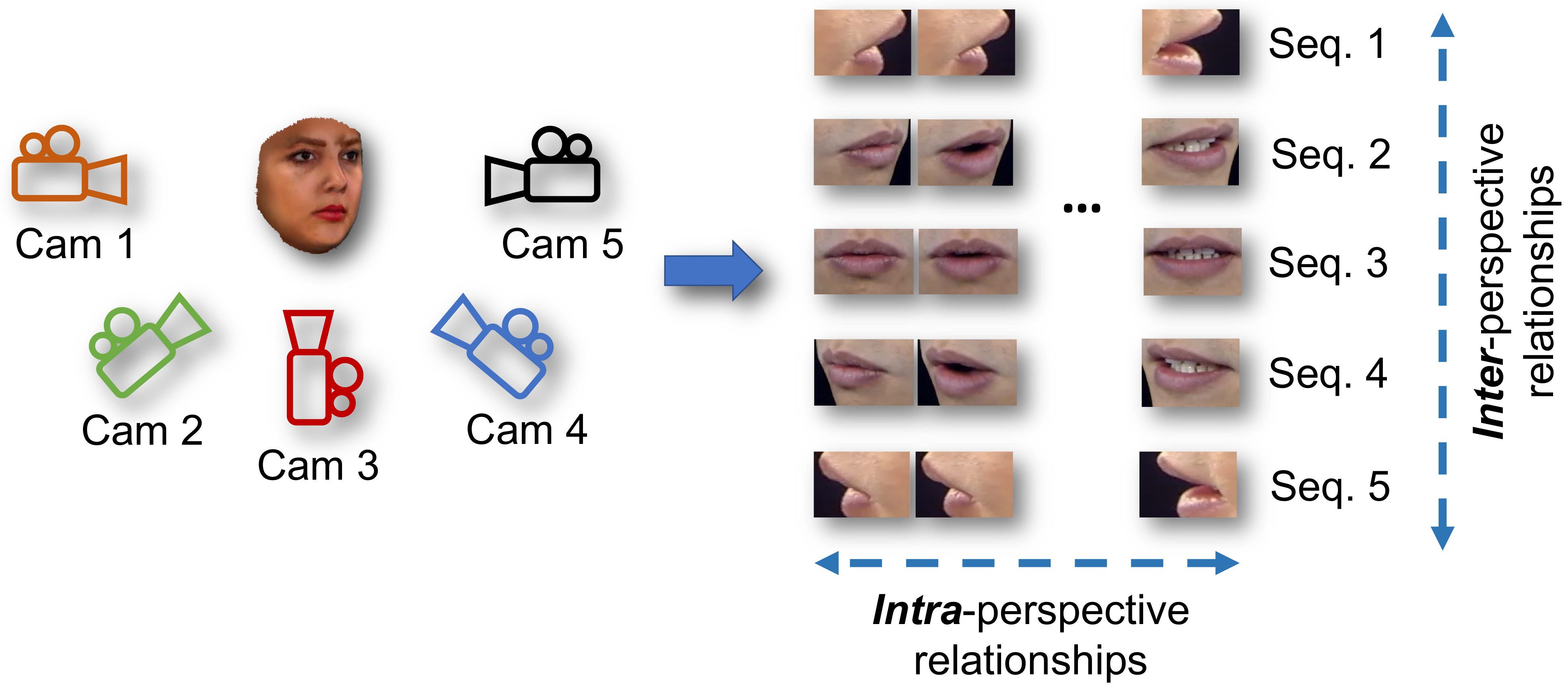}
    \end{center}
\caption{{Sequences captured from multiple perspectives include intra- and inter-perspective relationships that need to be effectively learned for robust visual recognition. We propose a novel LSTM cell capable of jointly learning incoming visual representations from various perspectives.}}
\label{fig:overview}
\end{figure}

Today, images and videos captured from multiple visual perspectives (multi-perspective) or view-points are extensively available thanks to the wide-spread adoption of consumer-level cameras, notably in smartphones, able to capture visual scenes simultaneously from multiple angles~\cite{CRV}. Multi-perspective sequences can be recorded by \textit{i}) several video cameras positioned at different angles, simultaneously acquiring the sequences, each of which including multiple samples/instances along time; and/or \textit{ii}) multi-view cameras such as Light Field (LF) cameras~\cite{LF}, from which all samples/instances of all sequences are acquired at a single time instant, e.g., with changing horizontal and vertical perspectives. We call these two types of multi-perspective sequences \textit{multi-perspective sequences over time} and \textit{multi-perspective sequences over space}, respectively. When either of these sequences are used for visual recognition tasks, it is possible to exploit both the \textit{intra-perspective} relationships (within each input/view sequence) and the \textit{inter-perspective} relationships (between the different input/view sequences), as illustrated in Figure \ref{fig:overview}.

Recurrent neural networks (RNN)~\cite{RNNSurvey} such as long short-term memory (LSTM)~\cite{LSTM}, have been widely used for learning sequential data. Nonetheless, conventional or \textit{vanilla} LSTM networks, hereafter referred to only as LSTM networks~\cite{LSTM, LSTMdesc}, learn from a single sequence, as each cell only accepts an instance of one particular sequence. In this context, in order to learn from multiple sequences (e.g., multi-perspective sequences), a separate LSTM network needs to be learned for each input sequence. As a result, inter-sequence relationships such as inter-perspective information are typically not learned. To aggregate the information learned by individual LSTM networks, fusion strategies have often been adopted~\cite{fuse2,fuse8,fuse10,CSVT}. Score-level fusion, also known as \textit{late fusion}, can be employed to combine the classification scores using different strategies such as a [weighted] sum rule or voting. This approach implies that the overall learning strategy is unable to learn the inter-sequence relationships and only relies on the aggregation of the class probabilities for the final decision. 
To avoid this problem, feature-level fusion, also known as \textit{early fusion}, can be used by concatenating the input sequences and feeding them consecutively to a single  network. Nonetheless, in this approach, the input representation is treated as a whole and not as different simultaneous perspectives captured from the same event, whereas in reality, different parts of this representation convey overlapping or complimentary information about the scene. Hence the parameters of the network are learned irrespective of the relationships between the available sequences, which are located in different parts of the concatenated representation. 

In this paper, we propose a novel Multi-Perspective LSTM (MP-LSTM) cell architecture to jointly learn the intra-perspective and inter-perspective relationships available in multi-perspective sequences. {To this end, we modify the conventional LSTM cell architecture, by incorporating additional gates and cell memories to adopt a novel recurrent joint learning strategy. These modifications enable our novel LSTM architecture to jointly update the long-term shared cell memory with respect to the information associated to several input perspective sequences simultaneously. This leads to more effective learning of the available inter-perspective relationships by identifying the complimentary or contradicting information across the perspective sequences when creating the output feature representations.} Our experiments show that the proposed MP-LSTM networks can learn richer representations to achieve better performance as exemplified for our experiments on two different visual recognition tasks.

The main contributions can be summarized as follows:
    (\textbf{1}) We propose the novel MP-LSTM cell architecture capable of jointly learning the intra- and inter-perspective relationships available in multi-perspective sequences;
    (\textbf{2}) we integrate our MP-LSTM network into two visual recognition solutions, for lip reading and face recognition tasks, covering two different types of multi-perspective sequences over time and over space;
    (\textbf{3}) our solutions achieve superior results over the state-of-the-art, with considerable performance gains of up to 5\% when multi-perspective information is jointly learned using our proposed model compared to other joint-learning or fusion strategies; and
    (\textbf{4}) we make our implementation publicly available\footnote{\href{https://github.com/arsm/MPLSTM}{https://github.com/arsm/MPLSTM}} to enable reproducibility and future comparisons.


\section{Related Work}

\subsection{Background}

LSTM networks are generally used to effectively learn long-term dependencies within a sequence~\cite{LSTMOD}. The LSTM cell architecture with peephole connections~\cite{peep} has been widely used for several learning tasks using sequential data~\cite{LSTMOD}. An LSTM network is composed of multiple LSTM cells, with a shared memory, called \textit{cell state}, to keep track of long-term dependencies over the network. This shared memory is controlled by an \textit{input} and a \textit{forget} gate, allowing the network to update the long-term memory considering the new incoming information. The updated cell state, along with an \textit{output gate}, then produce the output of the LSTM cell, known as the \textit{hidden state}. The networks created using these LSTM cells are often designed to take only one sequence as input. 

{Conventional approaches for dealing with multiple input sequences by LSTM networks use fusion strategies to either concatenate the input sequences and feed them consecutively to a single network (feature-level fusion) or combine the output scores obtained from independent LSTM networks applied to each input sequence (score-level fusion). Recently, there have been a few LSTM variants proposed to deal with multiple input sequences at the cell level~\cite{MVLSTM, LSTMact2, dualLSTM, joint}. In this context, novel architectures have been designed by adding, removing, modifying, or coupling gates, memory cells, or the connections between them, inside the LSTM cell. These novel architectures have been designed to allow each cell to jointly learn the intra-sequence (e.g., intra-perspective) along with inter-sequence (e.g., inter-perspective) dependencies across the input space.}

\subsection{Multi-Input LSTM Cell Architectures}
Since this paper's contribution is to propose a novel LSTM cell architecture for multi-perspective visual representation learning, we review the available multi-input LSTM cell architectures in the following. 
Multi-View LSTM (MV-LSMT)~\cite{MVLSTM} uses independent gates and cell states for each input sequence. The cell states corresponding to each input are first updated and are then concatenated to obtain the fused cell state. The fused cell state along with output gates form individual hidden states which are finally concatenated to produce the output and feed the next LSTM cell. It is worth mentioning that the ``Multi-View'' term used in~\cite{MVLSTM} does not imply different visual perspectives, as used in our paper. Instead, the term ``view'' is defined in generic terms, relating to the particular way of observing a phenomenon. This cell architecture was designed to fuse images and their text captions for image captioning tasks.
The Spatio-Temporal LSTM (ST-LSTM)~\cite{LSTMact2} cell architecture uses independent gates (except the output gate) for each of the two input sequences, independently updating the cell state for each input. A fused cell state, controlled by the output gate corresponding to the first input sequence, then produces the cell output. This cell architecture was designed to fuse RGB and human skeleton information for activity recognition.
The Dual-Sequence LSTM (DS-LSTM)~\cite{dualLSTM} cell architecture concatenates samples from two input sequences at a given instant to calculate the gates. Its gating functions are similar to the conventional LSTM, the difference being the way in which the candidate vectors are calculated. In this architecture, each input sequence independently contributes a candidate vector to be added to the cell state when updating the cell memory. This cell architecture was designed to simultaneously learn from two spectrogram sequences for speech recognition. 
The Gate-Level Fusion LSTM (GLF-LSTM)~\cite{joint} cell architecture considers a fusion scheme at the gate-level using independent forget, input, and output gates for each input sequence. The outputs of these gates are added to compute the fused values, thus determining the cell and hidden states. This cell architecture was designed for face recognition. 
The State-Level Fusion LSTM (SLF-LSTM)~\cite{joint} cell architecture considers fusion at the state-level, to learn the independent cell and hidden memory states from two simultaneous inputs for face recognition. These memory states are then added to produce the jointly learned outputs of the cell. 


Given that the cell state incorporates learnable parameters of the input and forget gates, and controls the output of the cell, this state can be a key component for jointly learning the inter-perspective relationships available in multi-perspective sequences. Nonetheless, a review of the related work (discussed above) indicates that in these works, the cell states are learned independently for each input sequence and are subsequently fused. This prevents the cell from identifying the complimentary or contradicting information between the multi-perspective sequences when producing the outputs. In our work, we address this shortcoming by introducing a new strategy to jointly learn the cell state. It should be noted that all the above-mentioned multi-input cell architectures will be considered for benchmarking when evaluating the proposed cell architecture and network.

\section{Proposed Method}

This section presents our novel MP-LSTM cell and network architectures to be used for representation learning.

\subsection{Model Overview}
Given multi-perspective sequences $S^p_{i}: p \in \{1,2,\dots, m\}, i \in \{1,2,\dots, n\}$, where $m$ is the number of simultaneously acquired perspectives and $n$ is the number of perspective instances, the MP-LSTM network $\mathcal{G}$ can be formulated as:
\begin{equation} \label{mvlstm}
\begin{gathered}
H_i=\mathcal{G}(S^p_{i}),  ~~~~~ \forall~p\in(1,...,m), \forall~{i}\in(1,...,n),
\end{gathered}
\end{equation}
where $H_i$ is the hidden state output by the $i^{th}$ cell. We use the term ``instance'' to be general enough to cover both \textit{i)} a moment in a multi-perspective sequence in time and \textit{ii)} a spatial location in a multi-perspective sequence in space.

It is a common practice in many visual recognition tasks to first extract spatial features from a given sequence $S^p_{i}$ prior to learning the sequential information through an RNN~\cite{LSTMdesc}. In this context, $S^p_{i}$ can be used as input to a feature extractor 
$\mathcal{X}$ such as a CNN, in order to first extract spatial features $E^p_{i}$, as formulated in Equation \ref{ext}. $S^p_{i}$ can then be substituted by $E^p_{i}$ in Equation \ref{mvlstm}.
\begin{equation} \label{ext}
\begin{gathered}
E^p_{i}=\mathcal{X}(S^p_{i}), 
~~~~~  \forall~p\in(1,...,m), \forall~{i}\in(1,...,n).
\end{gathered}
\end{equation}

\subsection{Cell Architecture}

Unlike the existing multi-input LSTM cell architectures, our proposed MP-LSTM cell jointly updates the cell state using the multi-perspective input sequences. To do so, additional gates and cell memories are incorporated into the cell architecture and a novel recurrent joint learning strategy is proposed that will be discussed in the following.

\begin{figure}[!t]
\centering
\includegraphics[width=0.85\columnwidth]{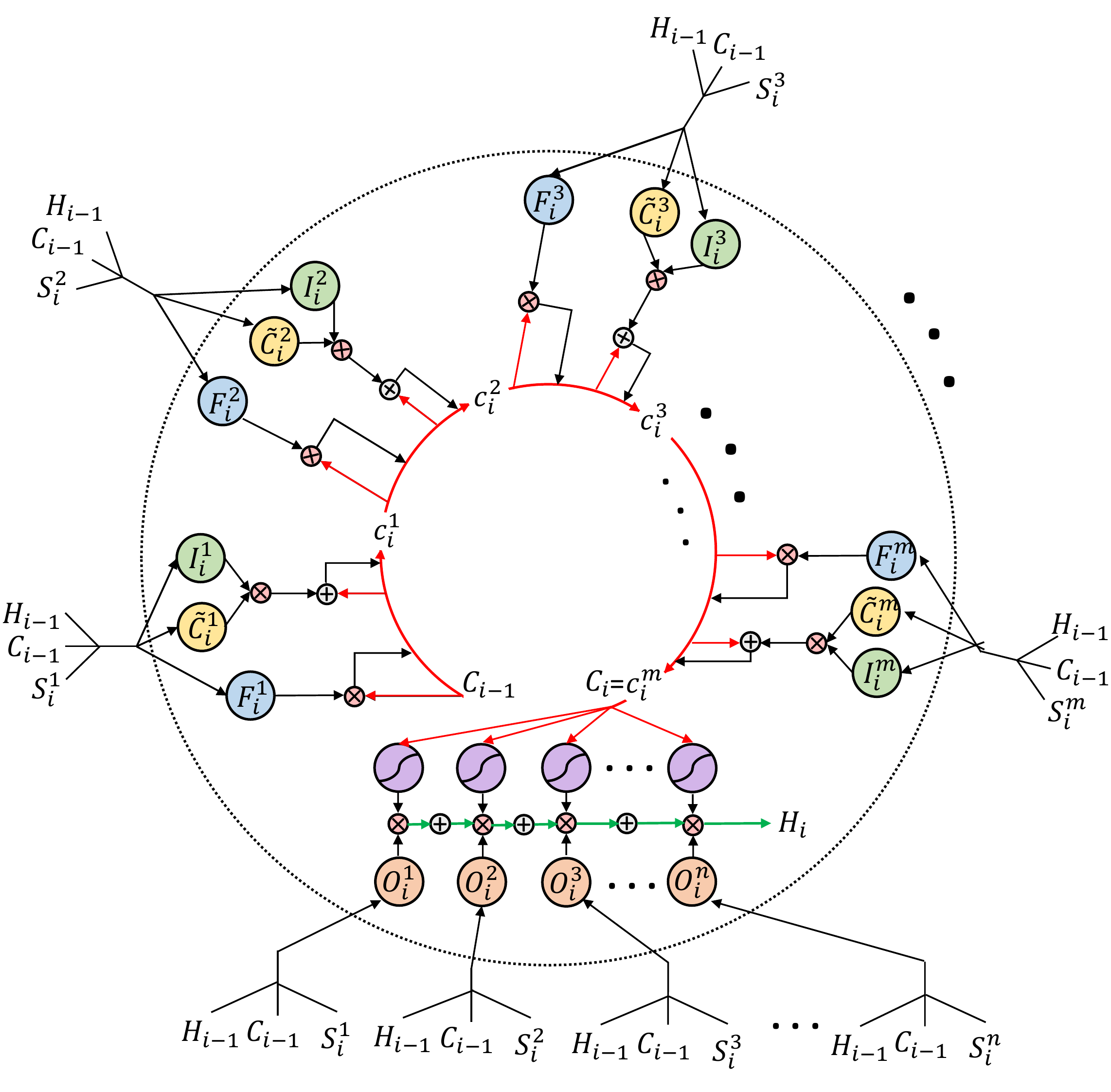}
\caption{Proposed MP-LSTM cell architecture.}
\label{fig:arch}
\end{figure}

The proposed cell architecture with peephole connections~\cite{peep} is illustrated in Figure \ref{fig:arch}. First, the input gates $I^p_i$, for the $i^{th}$ instance of the $p^{th}$ perspective sequence, are computed according to Equation \ref{input}, thus controlling the new information to be added to the shared cell state. The input gate inputs are the present sequence instance $S^p_i$, previous hidden state $H_{i-1}$, and previous cell state $C_{i-1}$:
\begin{equation} \label{input}
\begin{gathered}
I^p_i=\sigma(W^p_{Is} S^p_{i} + W^p_{Ih} H_{i-1}+W^p_{Ic} C_{i-1} +b^p_{I}),
\\  \forall~p\in(1,...,m),
\end{gathered}
\end{equation}
where $W^p_{Is}$, $W^p_{Ih}$, and $W^p_{Ic}$ are the input gate weights and $b^p_{I}$ is the input gate bias for the $i^{th}$ instance of the $p^{th}$ perspective sequence. $\sigma$ denotes the sigmoid activation, ensuring that the input value is bounded in the range [0,1].

\begin{figure*}[!t]
\centering
\includegraphics[width=0.75\linewidth]{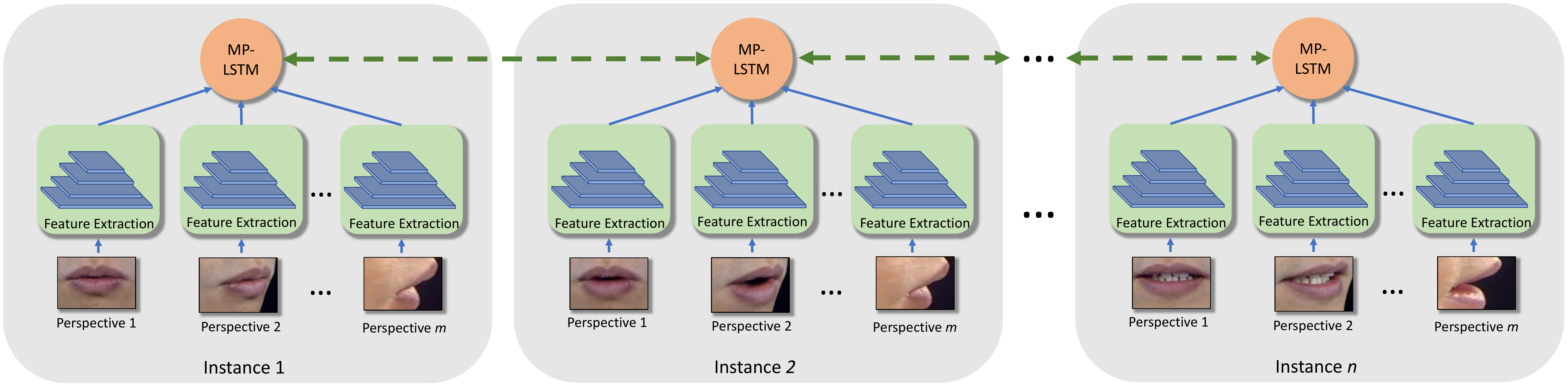}
\caption{The bi-directional MP-LSTM network architecture composed by our MP-LSTM cells where $n$ and $m$ are, respectively, the number of perspectives and associated instances.} 
\label{fig:net}
\end{figure*}

The vector of candidate values $\tilde{C}^p_i$, for the $i^{th}$ instance of the $p^{th}$ perspective sequence, is computed according to:
\begin{equation} \label{candidate}
\begin{gathered}
\tilde{C}^p_i=\tanh(W^p_{\tilde{C}s} S^p_{i} + W^p_{\tilde{C}h} H_{i-1}+W^p_{\tilde{C}c} C_{i-1} +b^p_{\tilde{C}}),
\\  \forall~p\in(1,...,m),
\end{gathered}
\end{equation}
where $W^p_{\tilde{C}s}$, $W^p_{\tilde{C}h}$, and $W^p_{\tilde{C}c}$ are the weights and $b^p_{\tilde{C}}$ is the bias for the vector of candidate values.
The $\tanh$ activation function is used to generate the output in the [-1,1] range while allowing for non-linearities to occur in the network. 
The vector of candidate values measured above holds the weights that can later be fully/partly added to the shared cell state with respect to the input gate.

In order to control how to forget perspective information from the shared cell state, the forget gate, $F^p_i$, for the $i^{th}$ instance of the $p^{th}$ perspective sequence, is computed:
\begin{equation}\label{forget}
\begin{gathered}
F^p_i=\sigma(W^p_{Fs} S^p_{i} + W^p_{Fh} H_{i-1}+W^p_{Fc} C_{i-1} +b^p_{F}),
\\  \forall~p\in(1,...,m),
\end{gathered}
\end{equation}
where $W^p_{Fs}$, $W^p_{Fh}$, and $W^p_{Fc}$ are the forget gate weights and $b^p_{I}$ is the forget gate bias.

Next, the cell state of the first sequence, $c^1_i$, is updated according to Equation \ref{C1}. This means that $c^1_i$ keeps the information coming from the previous cell, and the first perspective sequence observed at the current instance $i$. 
\begin{equation}\label{C1}
c^1_i= F^1_i \odot C_{i-1} + I^1_i \odot \tilde{C}^1_{i},
\end{equation}

The subsequent cell states of the other sequences, $c^p_i$: $\forall~p\in(2,...,m)$, are then updated using Equation \ref{Cp}. The cell state $c^p_i$ for the $p^{th}$ sequence is updated with respect to the jointly learnt cell state, $c^{p-1}_i$. This equation allows the cell to establish a relationship between the multi-perspective sequences, thus identifying complimentary or contradicting information to be learned or ignored.
\begin{equation}\label{Cp}
\begin{gathered}
c^p_i= F^p_i \odot c^{p-1}_i + I^p_i \odot \tilde{C}^p_{i};
~~~~~  \forall~p\in(2,...,m),
\end{gathered}
\end{equation}

The cell state that is updated using the last perspective sequence, $c^m_i$, is the new joint cell state, $C^i$, which includes the jointly learned information coming from all perspective sequences after the $i^{th}$ instance, formulated as $C_i= c^m_i$.

The jointly learned cell state can then be used to produce the output of the cell. To control how to update the hidden states of the perspective sequences, the output gates, $O^p_i$, are computed according to: 
\begin{equation}\label{output}
\begin{gathered}
O^p_i=\sigma(W^p_{Os} S^p_{i} + W^p_{Oh} H_{i-1}+W^p_{Oc} C_{i-1} +b^p_{O}),
\\ \forall~p\in(1,...,m),
\end{gathered}
\end{equation}
where $W^p_{Os}$, $W^p_{Oh}$, and $W^p_{Oc}$ are the output gate weights and $b^p_{O}$ is the output gate bias.

Each  perspective sequence's hidden state, $h^p_i$, is computed based on the jointly learnt cell state, $C_i$, and the output gates, according to:
\begin{equation}\label{hidden}
\begin{gathered}
h^p_i= O^p_i \odot \tanh C_i;
~~~~~  \forall~p\in(1,...,m),
\end{gathered}
\end{equation}

Finally, the output of the cell, $H_i$, across all perspectives, after the $i^{th}$ instance, is computed by adding the hidden states of each perspective as:
\begin{equation}\label{hiddenCell}
H_i= \sum_{p=1}^{m} h^p_i.
\end{equation}

\subsection{Network Architecture} 

The MP-LSTM cells can be connected to create a network capable of learning effective and richer representations for multi-perspective sequences. In this context, the output of the $i^{th}$ cell, $H_i$, as well as the jointly learned cell state, $C_i$, corresponding to $i^{th}$ input instances of all perspective sequences, are fed to the $({i+1})^{th}$ cell, respectively as short and long-term memories. The ${(i+1)}^{th}$ cell additionally receives the ${(i+1)}^{th}$ instances from all perspective sequences. This creates the network, as illustrated in Figure \ref{fig:net} (considering lip reading samples as example). The architecture presented in Figure \ref{fig:net} shows a \textit{bi-directional} network, as both forward and backward joint relationships are considered~\cite{bilstm} to form two feature vectors that are subsequently concatenated. Naturally, the proposed cell architecture can also be adopted in the context of other LSTM network architectures~\cite{LSTMOD}. In the experiments (Section 2.7), we will compare the performance of bi-directional and uni-directional network architectures. 

The number of cells in the network is equal to the number of instances available in each perspective sequence. The output of each cell takes into account the joint short- and long-term relationships observed up to that cell’s input. It should be noted that the network requires synchronized multi-perspective sequences to be received, with the same length (number of instances), as illustrated in Figure \ref{fig:net}. 
The network initializes the hidden and cell states to zero for the first cell. Depending on the learning task, other initialization mechanisms can be adopted for improving the learning performance or accelerating the training process~\cite{init}.

\section{Experiments and Performance Assessment}

In this section, we describe the recognition solutions and the experiments designed to demonstrate the effectiveness of the proposed MP-LSTM network.

\subsection{Experiment Setup}
We evaluate the applicability of the proposed network to visual representation learning on three public datasets for two different tasks, covering two different types of multi-perspective sequences: multi-perspective sequences over time (Experiment 1) and over space (Experiment 2). 

\noindent \textbf{Experiment 1 (Lip Reading):} In this experiment, we consider temporal sequences (videos) recorded from multiple perspectives using several video cameras capturing different angles over the scene. The MP-LSTM network is used to explore the inter-perspective dynamics over time for lip reading, also known as visual speech recognition.

\noindent \textbf{Experiment 2 (LF-Based Face Recognition):} In this experiment, we consider the usage of LF images, obtained by an LF camera, which simultaneously captures the intensity of light rays coming from multiple directions in space at a single time instant~\cite{LF}. LF cameras can provide multi-perspective sequences, for instance corresponding to the perspectives in the horizontal and vertical directions. In this experiment, the used LF images have been captured from 15$\times$15 different perspectives horizontally and vertically~\cite{CSVT}. The goal of this experiment is to evaluate the performance of the MP-LSTM network by exploiting the relationships between these two spatial perspective sequences for the face recognition task. 

\subsection{Recognition Solutions}
For both lip reading and face recognition experiments, we design solutions using our proposed MP-LSTM cell architecture. These solutions first use the ResNet-50 CNN~\cite{resnet}, pretrained on the large-scale VGG-Face2 dataset~\cite{csxpz18}, to extract representations from the each of the input sequences/perspectives. We additionally employ the 4-layer CNN proposed in~\cite{BMVC}, pretrained on the OuluVS2 dataset~\cite{lipoulu}, for lip reading. The extracted features are then fed to a bi-directional MP-LSTM network, as discussed in Section 3.3, followed by a soft-attention mechanism~\cite{b17} for selectively focusing on the most salient jointly learned states. A softmax classifier is finally used to perform the classification.

\subsection{Datasets and Test Protocols}
Here we describe the three datasets used in our experiments as well as the protocols used for evaluation purposes.

\begin{figure}[!h]
\centering
\includegraphics[width=0.61\columnwidth]{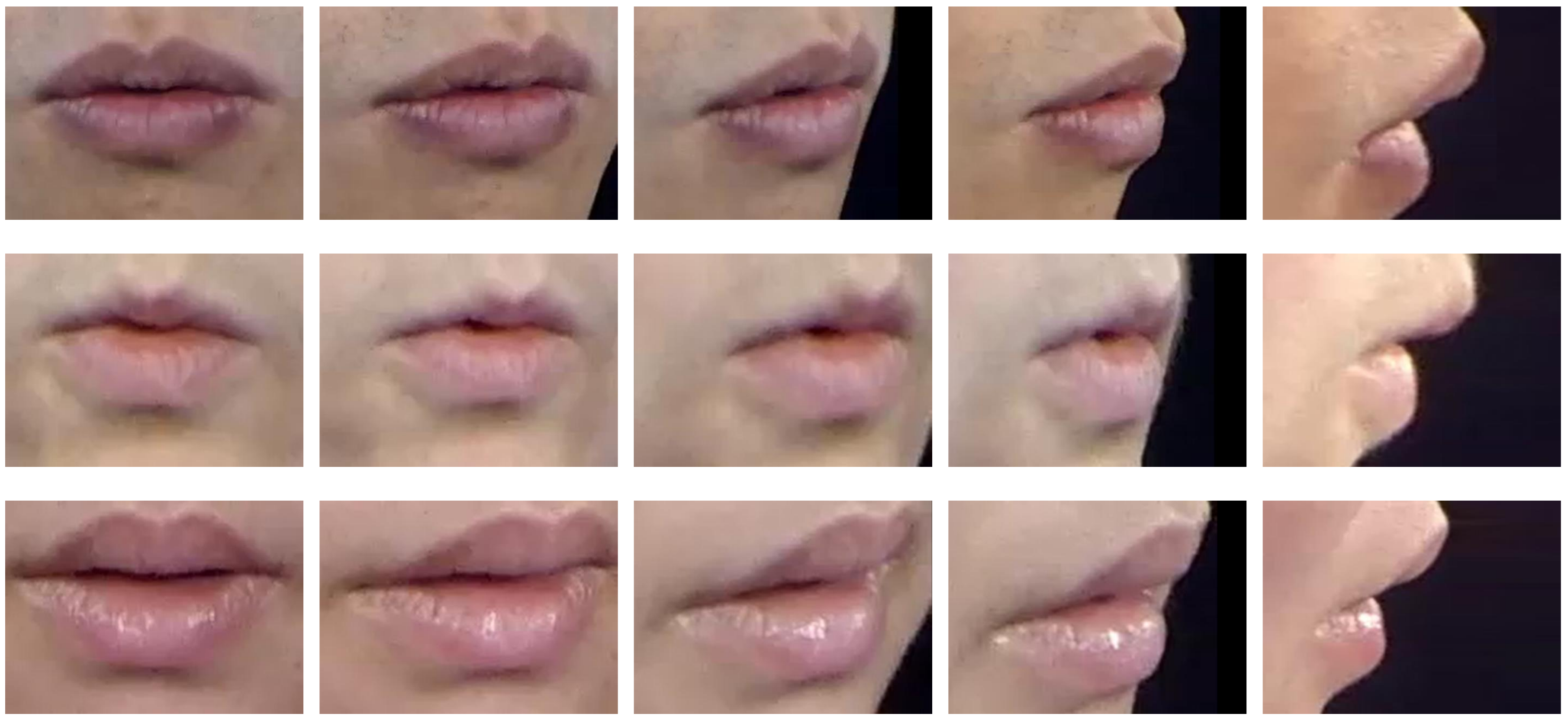}
\caption{Illustration of OuluVS2 samples from three different subjects captured from five different angles~\cite{lipoulu}.}
\label{fig:Oulu}
\end{figure}

\noindent \textbf{Lip Reading Dataset:} The OuluVS2 dataset~\cite{lipoulu} consists of 52 speakers uttering the same 10 phrases and 10 pre-determined digit sequences. The videos have been simultaneously recorded from five different viewing angles, spanning between the frontal and profile perspectives, as illustrated in Figure \ref{fig:Oulu}. In our experiments, we follow the test protocol proposed in~\cite{lipoulu}, conducting a speaker-independent experiment with 12 specified subjects for testing, 12 subjects for validation, and the remaining ones for training. We additionally employ the test protocol used in~\cite{BMVC,bilstm}, using only 10 phrases (output classes) for training and testing. 

\vspace{-2mm}

\begin{figure}[!h]
\centering
\includegraphics[width=0.78\columnwidth]{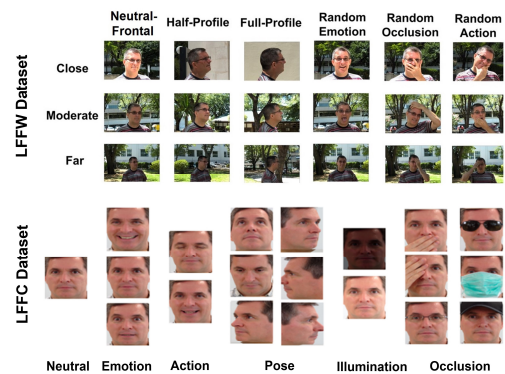}
\caption{Variations of a specific subject in the LFFW and LFFC datasets~\cite{TIP}.}
\label{fig:LFFC}
\end{figure}

\noindent \textbf{LF Face Datasets:} 
The LF Faces in the Wild (LFFW) and LF Face Constrained (LFFC) datasets~\cite{TIP} have been used in Experiment 2. LFFW includes 1908 LF face images, corresponding to 429,300 2D images, captured from 53 subjects under several unconstrained acquisition variations, in both indoor and outdoor environments. These LF images have been captured at different locations and from different distances, as illustrated in Figure \ref{fig:LFFC}. On the other hand, LFFC contains 1060 LF images, corresponding to 238,500 2D images, captured from the same 53 subjects used in the LFFW, but in a controlled acquisition setup. LFFC has been acquired between 1 day and 3 years prior to LFFW. The available images have different facial variations, including facial emotions, actions, poses, illuminations, and occlusions, as illustrated in Figure \ref{fig:LFFC}. For comprehensive evaluation, a cross-dataset test protocol between these two datasets has been considered.

\vspace{-3mm}

\subsection{Benchmarks}
\label{sec:bench}

We compare our MP-LSTM -based solutions to a number of relevant benchmarks for each experiment.
The selected benchmarks can be classified into three categories. 
\textit{i}) We replace the MP-LSTM cells in our solutions with conventional LSTM cells and instead use feature-level and score-level fusion strategies to obtain a single output given the different perspectives. 
\textit{ii}) We replace the MP-LSTM cell in our solutions with \textit{multi-input LSTMs}~\cite{MVLSTM, LSTMact2, dualLSTM, joint} reviewed in Section 2. This eliminates the need for a fusion step after learning from each perspective sequence. \textit{iii}) We also consider other alternative recognition solutions available in the literature for each experiment. For lip reading, we include PCA+LSTM+HMM~\cite{ACCV}, CNN+LSTM~\cite{ACCV2}, 4-layer CNN+Hierarchical LSTM~\cite{BMVC}, and VGG-M+Attentive Bi-LSTM~\cite{atlstm}, as they use LSTM networks in combination with different spatial feature extractors and classifiers. 3DCNN~\cite{ACCV2} has also been considered as it has shown to be an effective alternative to CNN + LSTM architectures. It is worth noting that there are some lip reading solutions that use other pre-processing or post-processing steps, particularly using \textit{both audio and visual} information~\cite{LipAAAI}, to boost the recognition performance. These solutions have not been included here in order to perform a fair comparison, as the main goal here is to show the effectiveness of our MP-LSTM proposal, notably in comparison with other fusion strategies and the existing multi-input LSTM learning architectures. For face recognition, two high performance CNNs including ResNet-50 and squeeze-and-excitation (SE) ResNet-50 pretrained on the large-scale VGG-Face2 dataset~\cite{csxpz18}, as well as the state-of-the-art VGG-16 + LSTM~\cite{CSVT} solution are selected. Naturally, the proposed MP-LSTM cell can be adopted as part of other multi-perspective visual recognition solutions that include an LSTM module while considering other solutions for face recognition and lip reading~\cite{LIP3}.

\subsection{Implementation Details}
The optimal parameters for achieving the best performance results for the face recognition and lip reading experiments are summarized in Table \ref{tab: parameter}. This table includes the best hyperparameter values, empirically obtained for each of the sub-networks, notably the CNN feature extractor, the MP-LSTM network, and the attention mechanism, along with the parameters used to train the end-to-end network as a whole. The metadata available with the three datasets was used to crop the face and mouth regions from the original images for the face recognition and lip reading experiments, respectively. The implementation was done using Keras~\cite{keras} with TensorFlow backend~\cite{tensorflow}, and the training used an Nvidia GeForce GTX 1080 Ti GPU.

\begin{table}[!t]
\centering
\caption{Best parameter values empirically obtained for the face recognition and lip reading solutions.}
\setlength
\tabcolsep{3.5pt}
\scriptsize
\begin{tabular}{ l| l| l| l}
\hline
\textbf{Sub-Net.} & \textbf{Parameter} & \textbf{Face Rec.}& \textbf{Lip Reading} \\
\hline
\hline
    CNN   & Architecture  & ResNet-50 & ResNet-50 \& 4-Layer CNN  \\
    Feature & Pretrained  & VGG-Face2 & VGG-Face2 \& OuluVS2 \\
    Extractor  & \# of Inputs  & 15 $\times$ 2 & 200 $\times$ \# of Seq.  \\
      & Embedding Layer  & Avg. Pooling & Avg. Pooling \& Last Layer \\
      & Feature Size  & 2048 & 2048 \& 450 \\
    \hline

     MP-LSTM & \# of Inputs  & 15 $\times$ 2 & 200 $\times$ \# of Seq. \\
      & \# of Outputs  & 15 & 200\\
      & Hidden Size  & 256 $\times$ 2  & 128 $\times$ 2\\
      & Dropout Rate  & 0.1  & 0.1\\
      & Network Arch.  & Bi-directional & Bi-directional\\
    \hline

    Attention & Activation Func.  & Softmax & Softmax\\
    \hline    

     Full Network  & Batch Size & 53 & 120 \\
      & Loss Function & Cross-entropy & Cross-entropy\\
     & Optimizer & Rmsprop & Rmsprop\\
     & Metric & Accuracy & Accuracy\\
     & \# of Epochs & 100 & 200\\
    \hline
\end{tabular}
\label{tab: parameter}
\end{table}

\begin{figure}[!h]
\centering
\includegraphics[width=1\columnwidth]{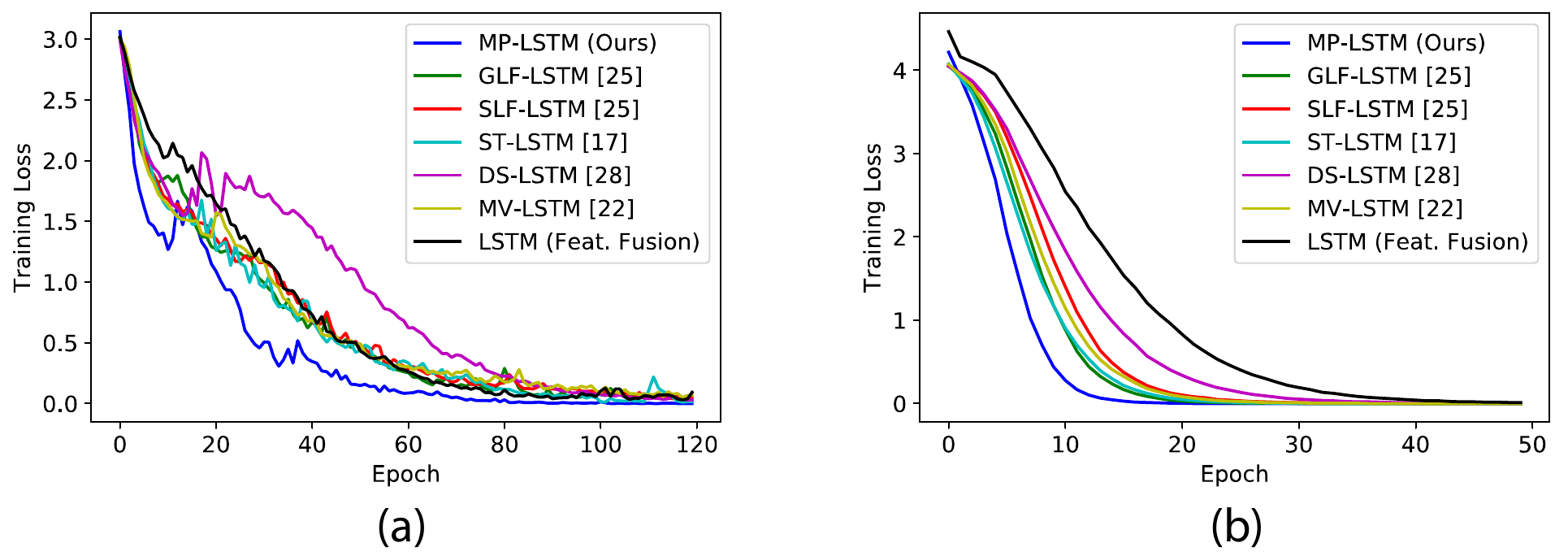}
\caption{{Training losses for our proposed LSTM network and several alternative multi-input LSTM variants for (a) lip reading; and (b) face recognition experiments.}}
\label{fig:loss}
\end{figure}

{\subsection{Training}
The training losses for our MP-LSTM network are plotted in Figure \ref{fig:loss} for the lip reading and face recognition experiments, along with results for the other multi-input LSTMs discussed in Section 2. It can be observed that training for face recognition is smoother and faster than for lip reading; this may be associated to the smaller size of the datasets, fewer number of perspectives, and narrower angular information. Nevertheless, the results clearly show that our network converges faster than the other LSTM variants for both tasks/experiments.}

\subsection{Performance}

For the lip reading experiment, we selected three perspectives, including the frontal ($0^{\circ}$), half-profile ($45^{\circ}$), and full-profile ($90^{\circ}$) perspectives, while the $30^{\circ}$ and $60^{\circ}$ perspectives have been omitted due to space constraints in our paper. In this context, all 2-perspective combinations as well as 3-perspective combination have been used as inputs to our proposed network. For the LF-based face recognition evaluation, horizontal and vertical perspective sequences were selected as the two inputs to the proposed network.

\begin{table}[!t]
\centering
\caption{Experiment 1: Comparison of MP-LSTM with alternative multi-perspective lip reading solutions using all classes.}
\setlength
\tabcolsep{3pt}
\scriptsize
\begin{tabular}{  l| l| l| l| l }
\hline
    \multicolumn{1}{ c |}{\textbf{Solution}} & \multicolumn{4}{ c }{\textbf{Perspective Angle }}\\ 
    \hline
     & \textbf{$0^{\circ}$-$45^{\circ}$} & \textbf{$0^{\circ}$-$90^{\circ}$} & \textbf{$45^{\circ}$-$90^{\circ}$}& $0^{\circ}$-$45^{\circ}$-$90^{\circ}$\\ 
    \hline\hline

    ResNet-50 + LSTM (Feat. Fusion)  & 79.58\% & 78.61\%& 76.11\% & 80.69\% \\
    ResNet-50 + LSTM (Sco. Fusion)    & 77.78\% & 78.19\% & 72.23\% & 81.11\%\\
    \hline
    ResNet-50 + MV-LSTM ~\cite{MVLSTM}  &   78.61\%  &  80.55\% & 77.78\% & 81.94\%  \\
    ResNet-50 + ST-LSTM ~\cite{LSTMact2}   & 76.11\% & 80.27\% & 74.16\% & 78.47\%\\
    ResNet-50 + GLF-LSTM~\cite{joint} &   78.05\%  & 79.44\%  & 78.33\% & 79.16\% \\
    ResNet-50 + SLF-LSTM~\cite{joint} & 79.17\% & 79.86\%  &  79.58\% & 82.22\% \\
    ResNet-50 + DS-LSTM ~\cite{dualLSTM} &  77.50\% &  79.72\% & 76.11\% & 80.97\%\\
    \hline
    PCA+LSTM+HMM~\cite{ACCV}   &  73.90\% & 72.70\% & --- & --- \\
    3DCNN~\cite{ACCV2}   &  --- & --- & --- & 76.10\%\\
    CNN+LSTM~\cite{ACCV2} &  --- & --- & --- & 80.00\%\\
    \hline
    \textbf{ResNet-50 + MP-LSTM (Ours)}  &  \textbf{83.74\%} & \textbf{83.05\%} & \textbf{82.36\%} & \textbf{87.22\%}  \\

    \hline
\end{tabular}
\label{tab: LIPRes}
\end{table}

\begin{table}[!t]
\centering
\caption{Experiment 1: Comparison of MP-LSTM with alternative multi-perspective lip reading solutions using 10 classes.}
\setlength
\tabcolsep{2.pt}
\scriptsize
\begin{tabular}{  l| l| l| l| l }
\hline
    \multicolumn{1}{ c |}{\textbf{Solution}} & \multicolumn{4}{ c }{\textbf{Perspective Angle }}\\ 
    \hline
     & \textbf{$0^{\circ}$-$45^{\circ}$} & \textbf{$0^{\circ}$-$90^{\circ}$} & \textbf{$45^{\circ}$-$90^{\circ}$}& $0^{\circ}$-$45^{\circ}$-$90^{\circ}$\\ 
    \hline\hline
    VGG-M + Attentive Bi-LSTM~\cite{atlstm}   &  --- & --- & --- & 87.0\%\\
    4-Layer CNN + Hierarch. LSTM~\cite{BMVC} &  93.6\% & 94.8\% & 93.6\% & 95.6\%\\
    \hline
    \textbf{4-Layer CNN + MP-LSTM (Ours)}  &  \textbf{94.6\%} & \textbf{95.6\%} & \textbf{95.0\%} & \textbf{96.8\%}  \\

    \hline
\end{tabular}
\label{tab: LIPRes2}
\end{table}

\begin{table*}[!t]
\centering
\setlength\tabcolsep{3.5pt}
\scriptsize
\centering
\caption{Experiment 2: Comparison of MP-LSTM with alternative multi-perspective lip reading solutions.}
\begin{tabular}{l|llllll|llllll|l}
\hline
 \multicolumn{1}{ c |}{\textbf{Solution}}& \multicolumn{6}{ c }{\textbf{ Protocol 1 (Train: \textit{LFFC}, Test: \textit{LFFW}})} & \multicolumn{6}{ |c }{\textbf{Protocol 2 (Train: \textit{LFFW}, Test: \textit{LFFC}})} & \multicolumn{1}{ |c }{\textbf{Avg.}}  \\ 
\hline
 & \textbf{Neutral} & \textbf{ Exp.} & \textbf{H-Prof.} & \textbf{F-Prof.} & \textbf{Occl.} & \textbf{Act.}& \textbf{Neutral} & \textbf{ Exp.} & \textbf{Act.} & \textbf{Pose} & \textbf{Illum.} & \textbf{Occl.}  \\
\hline\hline
ResNet-50 + LSTM (Hor. seq.)  & 92.14\% & 88.99\% &  80.82\% &  58.18\% &  80.50\% &  71.70\% & 96.23\% & 95.60\% &  \textbf{94.34\%} &  91.51\% &  96.23\% &  83.33\% &  84.43\% \\
ResNet-50 + LSTM (Ver. seq.)  & 92.14\% & 88.99\% &  80.19\% &  59.75\% &  80.82\% &  71.38\% & 96.23\% & 94.97\% &  93.40\% &  92.45\% &  97.17\% &  82.39\% &  84.72\% \\
ResNet-50 + LSTM (Feat. Fus.)  & 92.45\% & 90.57\% &  80.50\% &  60.69\% &  82.08\% &  73.90\% & 96.23\% & 95.60\% &  93.40\% &  93.40\% &  96.23\% &  83.01\% & 85.53\% \\
ResNet-50 + LSTM (Sco. Fus.) & 92.77\% & 90.88\% &  80.19\% &  61.01\% &  82.70\% &  72.33\% & 96.23\% & 96.23\% &  93.40\% &  92.77\% &  97.17\% &  83.33\% & 85.55\% \\ 
\hline
ResNet-50 + MV-LSTM ~\cite{MVLSTM} & 92.45\% & 90.25\% &  \textbf{83.02\%} &  \textbf{61.64\%} &  80.19\% &  72.64\% & 96.23\% & 96.23\% &  \textbf{93.40\%} &  94.34\% &  98.11\% &  83.33\% & 85.86\% \\
ResNet-50 + ST-LSTM ~\cite{LSTMact2} & 93.39\% & 90.57\% &  81.76\% &  \textbf{61.64\%} &  82.07\% &  73.90\% & 98.11\% & 96.23\% &  92.45\% &  94.65\% &  97.17\% &  82.28\% & 86.27\% \\
ResNet-50 + GLF-LSTM~\cite{joint}   & 93.08\% & 90.25\% &  82.08\% &  60.06\% &  82.08\% &  73.58\% & 96.23\% & 96.23\% &  92.45\% &  93.08\% &  97.17\% &  85.53\% & 85.99\% \\
ResNet-50 + SLF-LSTM~\cite{joint} & 93.08\% & 89.31\% &  82.70\% &  60.69\% &  83.02\% &  73.27\% & 98.11\% & \textbf{97.48\%} &  \textbf{93.40\%} &  94.03\% &  \textbf{98.11\%} &  83.65\% & 86.16\% \\
ResNet-50 + DS-LSTM ~\cite{dualLSTM}   & 92.76\% & 89.62\% &  79.87\% & 60.37\% &  80.81\% &  73.58\% & 96.27\% & 96.86\% & \textbf{93.40\%} &  93.71\% &  97.17\% &  82.70\% & 85.51\% \\
\hline
SE-ResNet-50~\cite{csxpz18} & 78.30\%  & 77.35\% & 69.18\% & 51.88\% & 66.03\% & 58.80\% &  83.01\% & 79.87\% & 77.35\% & 81.76\% & 85.84\% & 69.81\% & 72.42\% \\
ResNet-50~\cite{csxpz18} & 87.73\%  &  86.16\%  & 77.67\%  & 52.51\% & 73.27\% & 67.29\% & 96.22\%  & 94.96\% & 92.45\% & 86.62\% & 96.22\% & 82.38\% & 81.81\%   \\
VGG-16 + LSTM~\cite{CSVT}  & 90.57\% & 86.16\% & 71.07\% & 34.48\% & 72.01\% & 60.69\% & 92.45\% & 89.94\% & 90.57\% & 76.10\% & 88.68\% & 73.27\% & 75.17\% \\
\hline
\textbf{ResNet-50 + MP-LSTM (Ours)} &\textbf{94.34\%} &  \textbf{91.19\%} & 82.39\% &  61.01\% &  \textbf{83.64\%} &  \textbf{75.47\%} & \textbf{100.00\%} & 96.86\% &  \textbf{93.40\%} &  \textbf{96.54\%} &  \textbf{98.11\%} &  \textbf{84.59\%} &  \textbf{87.18\%} \\
\hline
\end{tabular}
\label{tab:DB1}
\end{table*}

\noindent \textbf{Experiment 1 (Lip Reading):}
The lip reading performance results obtained by ResNet-50 + LSTM applied to the individual $0^{\circ}$, $45^{\circ}$, and $90^{\circ}$ perspectives (single-perspective) are respectively 74.86\%, 72.63\%, and 69.44\%.  
Table \ref{tab: LIPRes} presents the lip reading performance of our novel solution as well as the benchmarks presented in Section 4.4 using all the 20 classes. For lip reading, our solution outperforms all the other methods for all the viewing angle combinations by a large margin. Concerning the 2-perspective combinations, the best performing benchmark, SLF-LSTM~\cite{joint}, delivers the average performance results of 79.54\%, while MP-LSTM achieves an average additional gain of 3.51\%. This performance gain is increased to 5\% when considering the combination of all three perspectives, showing the superiority of our MP-LSTM network in the joint learning of inter-perspective relationships, while also converging faster than other multi-input LSTM variants (see Figure \ref{fig:loss}). Interestingly, some of the benchmarks such as ST-LSTM ~\cite{LSTMact2} and GLF-LSTM~\cite{joint} do not necessarily improve the results when the number of perspectives is increased from 2 to 3. However, a clear performance boost is observed with our solution, showing the ability of the MP-LSTM network in jointly learning from multiple perspectives. We additionally perform lip reading experiments using 10 phrases similar to~\cite{BMVC,bilstm}. To this end, we use the same 4-layer CNN used in~\cite{BMVC}, thus solely comparing our MP-LSTM network with the hierarchical LSTM network proposed in~\cite{BMVC}. The results show the superiority of our MP-LSTM when compared to attentive Bi-LSTM~\cite{atlstm} and hierarchical LSTM~\cite{BMVC}.

\noindent \textbf{Experiment 2 (LF-based Face Recognition):} Table \ref{tab:DB1} presents the face recognition performance when, respectively: \textit{i}) LFFC is used for training and LFFW for testing (Protocol 1); and \textit{ii}) LFFW is used for training and LFFC for testing  (Protocol 2). These tables present results for our proposed recognition solution that adopts the proposed MP-LSTM network, as well as for the benchmarks listed in Section 4.4. The performance results clearly show the added value of the fusion strategies for face recognition, when compared to the individual results, i.e. using only horizontal or vertical sequences. The results for the multi-input LSTMs benchmarks (rows 5-9) are generally better than the fusion-based solutions (rows 3 and 4), due to the joint exploitation of multi-perspective sequences. From the available multi-input LSTMs, ST-LSTM~\cite{LSTMact2} and SLF-LSTM~\cite{joint} perform better than the other variants. Finally, the results show that MP-LSTM achieves better performance than all of the available solutions, for most the test variations considered. 

{\noindent \textbf{Comparison to Gated Recurrent Unit (GRU):} We also adopt the same joint learning strategy for GRU~\cite{gru} compared to MP-LSTM. The results demonstrate a slight superiority for MP-LSTM over the multi-perspective version of GRU, respectively by achieving performance gains of 0.7\% and 0.9\% for lip reading and face recognition.}

{\noindent \textbf{Uni-Directional Vs. Bi-Directional MP-LSTM:} Table \ref{tab:bi} presents the recognition performance for uni-directional and bi-directional networks adopting our proposed cell architecture. The results show that a bi-directional network always achieves superior performance for both lip reading and face recognition tasks since both forward and backward relationships are considered. The performance gain is more evident for the most challenging case of lip reading, as it involves the combination of all possible perspectives.}

\begin{table}[]
\centering
\setlength\tabcolsep{2 pt}
\footnotesize
\centering
\caption{{{Recognition performance for uni-directional and bi-directional MP-LSTM networks.}}}
\begin{tabular}{l|c|c|c|c|c|c}
\hline

    \multicolumn{1}{ c |}{\textbf{Experiment}} & \multicolumn{4}{ c |}{\textbf{Lip Reading }} & \multicolumn{2}{ c }{\textbf{Face Rec. }}\\ 
    \hline
     \textbf{Protocol} & \textbf{$0^{\circ}$-$45^{\circ}$} & \textbf{$0^{\circ}$-$90^{\circ}$} & \textbf{$45^{\circ}$-$90^{\circ}$}& $0^{\circ}$-$45^{\circ}$-$90^{\circ}$ & Prot. 1 & Prot. 2\\ 
    \hline\hline

\textbf{Uni-Direc.}   & 82.92\%  &  80.42\% & 80.56\%  & 82.49\%  &  77.94\% & 90.00\% \\
\textbf{Bi-Direc.}   & 83.74\% &  83.05\%  & 82.36\%  & 87.22\%  &  81.34\%  & 93.02\% \\
\hline
\end{tabular}
\label{tab:bi}
\end{table}

\noindent \textbf{Impact of Camera Baseline on Performance:} The results in Tables \ref{tab: LIPRes} and \ref{tab:DB1} indicate that the performance gains obtained by adopting MP-LSTM are more significant for the lip reading experiment. This is likely due to the fact that the camera baseline (distance between the lenses/cameras) for the used LF camera, Lytro Illum~\cite{Lytro}, is very narrow. In this context, the perspective images are rendered for very close horizontal and vertical positions, i.e. a short baseline, implying there is less angular information to be learned from the LF images. In contrast, the lip reading videos are captured from ($0^{\circ}$), ($45^{\circ}$), and ($90^{\circ}$) angles, providing the MP-LSTM network with much wider angular information, i.e. a larger baseline, thus allowing it to learn richer joint representations and achieve better performance.

\subsection{Feature Space Exploration}
\begin{figure}[t!]
\centering
\includegraphics[width=1\columnwidth]{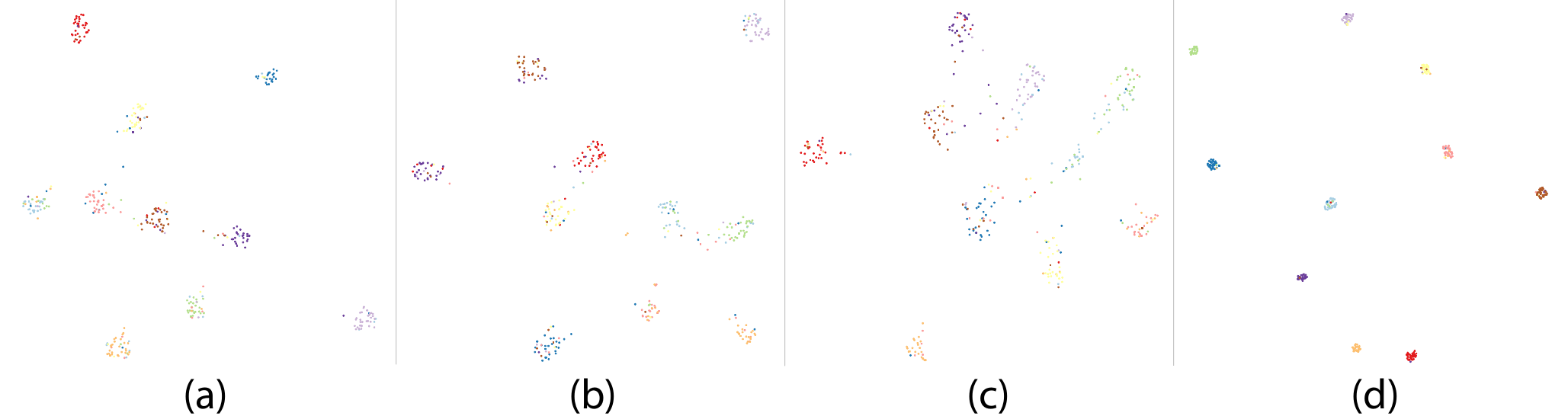}
\caption{UMAP visualization of the lip reading feature spaces produced by individual a) $0^{\circ}$; b) $45^{\circ}$; and c) $90^{\circ}$ perspectives, and when these three sequences are jointly learned by d) our solution. The colors denote the first 10 classes available in the OuluVS2 dataset.}
\label{fig:UMAP}
\end{figure}

We visualize the discriminative behaviour of the proposed MP-LSTM network using Uniform Manifold Approximation and Projection (UMAP)~\cite{umap}. Figure \ref{fig:UMAP} plots the feature spaces produced by several lip reading solutions in a two dimensional space using UMAP. Figure \ref{fig:UMAP} includes the UMAP plots when using the ResNet50+LSTM solution applied to the individual $0^{\circ}$ (Figure \ref{fig:UMAP}-a), $45^{\circ}$ (Figure \ref{fig:UMAP}-b), and $90^{\circ}$ (Figure \ref{fig:UMAP}-c) perspective sequences. The Figure \ref{fig:UMAP}-d includes the UMAP plot when combining these three sequences, using our MP-LSTM network. This visualisation is performed for the first 10 classes available in OuluVS2 dataset. A better representation should create denser clusters with less data points distributed far from their respective cluster's centroid to facilitate more accurate discrimination between the various classes. As Figure \ref{fig:UMAP} shows, our proposed MP-LSTM clearly results in more separable classes versus the other solutions.

{\subsection{Ablation and Configuration}}

We perform ablation experiments to understand how each part of our method contributes to the performance.
We design these experiments using all the three perspectives for the lip reading task using the 4-layer CNN to extract representations. 
We create three variants of the MP-LSTM cell, named Models A, B, and C.
In \emph{Model A} 
we remove the previous cell state, $C_{i-1}$, when computing the new cell state in Equation \ref{C1}. In \emph{Model B} we update the cell state in Equation \ref{Cp} with $c^{p}_{i-1}$ (the same perspective from the previous instance) instead of $c^{p-1}_{i}$ (the previous perspective from the same instance). In \emph{Model C} we remove the long-term memory coming from the previous cell. Finally, the complete model is referred to as \emph{Full Model}. The results presented in Table~\ref{tab: abl} demonstrate the superiority of the complete model when compared to the various reduced models.  

\subsection{Time Analysis}

Finally, we study the computational time for our LSTM network along with the other multi-input LSTM variants. This analysis has been done by measuring the training/testing times on a 64-bit Intel PC with a 3.20~GHz Core i7 processor, 48~GB RAM, and an Nvidia GeForce GTX~1080 Ti GPU, running TensorFlow with Keras backend. Table \ref{tab:time} shows the training times (in seconds) for each sequence when considering a combination of all the three perspectives for the lip reading. It should be noted that these times are only presented for the LSTM components of the entire model and do not include the time needed for training the CNN component. It can be observed from Table \ref{tab:time} that the required training time for our MP-LSTM network is less than that for the other LSTM variants, due to the faster convergence of our method (see Figure \ref{fig:loss}). Concerning testing, we observe that the time is very similar for all the methods, notably around  0.05 $\pm$ 0.01 seconds per sequence.



\begin{table}[!t]
\centering
\caption{Ablation study for the lip reading experiments.}
\setlength
\tabcolsep{5pt}
\footnotesize
\begin{tabular}{  l| l| l| l| l }
\hline
    Configuration & \textbf{Model A} & \textbf{Model B} & \textbf{Model C}& \textbf{Full Model}\\ 
    \hline
    Performance  &  { 93.1\%} & { 94.8\%} & { 90.9\%} & \textbf{ 96.8\%}  \\
    \hline
\end{tabular}
\label{tab: abl}
\end{table}

\begin{table}[]
\centering
\setlength\tabcolsep{2.5pt}
\footnotesize
\centering
\caption{{Average lip reading training time for our proposed and alternative multi-input LSTM networks (in seconds).}}
\begin{tabular}{l|c|c|c|c|c|c}
\hline
\textbf{Method}   & MV~\cite{MVLSTM}& ST~\cite{LSTMact2}& GLF~\cite{joint}& SLF~\cite{joint}& DS~\cite{dualLSTM}  & MP (Ours)  \\
\hline
\textbf{Time}   & 2.31 & 1.54 & 1.92 & 2.19 & 2.64 & \textbf{1.24} \\
\hline
\end{tabular}
\label{tab:time}
\end{table}

\section{Conclusion}

In this paper, we propose the Multi-Perspective LSTM (MP-LSTM) cell architecture for effectively learning multi-perspective sequences. Our approach exploits both the intra-perspective relationships within each view, as well as the inter-perspective dynamics over time or over space, by including additional gates and cell memories with respect to the conventional LSTM cell architecture to adopt a novel recurrent joint learning strategy. The performance of the MP-LSTM network is assessed in the context of two multi-perspective visual recognition tasks, namely lip reading and face recognition. The performance results clearly show the superior performance of our novel solutions over a large number of relevant benchmarks. The improvements are more evident when dealing with lip reading data, since the multiple sequences result from a larger camera baselines. 


{\small
\bibliographystyle{ieee_fullname}
\bibliography{references_new}

\begin{thebibliography}{10}\itemsep=-1pt

\bibitem{tensorflow}
M. Abadi, P. Barham, and J. Chen.
\newblock Tensorflow: A system for large-scale machine learning.
\newblock In {\em Symp. on Operating Systems Design and Implementation}, pages
  265--283, 2016.

\bibitem{lipoulu}
I. {Anina}, Z. {Zhou}, G. {Zhao}, and M. {Pietikäinen}.
\newblock {OuluVS2}: A multi-view audiovisual database for non-rigid mouth
  motion analysis.
\newblock In {\em Conference on Automatic Face and Gesture Recognition}, pages
  1--5, 2015.

\bibitem{csxpz18}
Q. Cao, L. Shen, W. Xie, M. Parkhi, and A. Zisserman.
\newblock {VGGFace2}: A dataset for recognising faces across pose and age.
\newblock In {\em Conference on Automatic Face and Gesture Recognition}, pages
  67--74, 2018.

\bibitem{gru}
K. Cho, B. Merrienboer, C. Gulcehre, D. Bahdanau, F. Bougares, H. Schwenk, and
  Y. Bengio.
\newblock Learning phrase representations using {RNN} encoder-decoder for
  statistical machine translation.
\newblock In {\em Conference on Empirical Methods in Natural Language
  Processing}, page 1724–1734, 2014.

\bibitem{keras}
Fran\c{c}ois Chollet.
\newblock Keras.
\newblock \url{https://keras.io}, 2015.

\bibitem{fuse10}
Q. Cui, S. Wu, Q. Liu, W. Zhong, and L. Wang.
\newblock {MV-RNN}: a multi-view recurrent neural network for sequential
  recommendation.
\newblock {\em IEEE Transactions on Knowledge and Data Engineering},
  32(2):317--331, 2018.

\bibitem{LSTMdesc}
J. {Donahue}, L.~A. {Hendricks}, M. {Rohrbach}, S. {Venugopalan}, S.
  {Guadarrama}, K. {Saenko}, and T. {Darrell}.
\newblock Long-term recurrent convolutional networks for visual recognition and
  description.
\newblock {\em IEEE Transactions on Pattern Analysis and Machine Intelligence},
  39(4):677--691, 2017.

\bibitem{fuse2}
H. Gammulle, S. Denman, S. Sridharan, and C. Fookes.
\newblock Two stream {LSTM}: A deep fusion framework for human action
  recognition.
\newblock In {\em Winter Conference on Applications of Computer Vision}, pages
  177--186, 2017.

\bibitem{peep}
F.~A. {Gers} and J. {Schmidhuber}.
\newblock Recurrent nets that time and count.
\newblock In {\em International Joint Conference on Neural Networks}, pages
  189--194, 2000.

\bibitem{bilstm}
A. Graves and J. Schmidhuber.
\newblock Framewise phoneme classification with bidirectional {LSTM} and other
  neural network architectures.
\newblock {\em Neural Networks}, 18(5):602--610, 2005.

\bibitem{LSTMOD}
K. Greff, R. Srivastava, J. Koutn{\'\i}k, B. Steunebrink, and J. Schmidhuber.
\newblock {LSTM}: A search space odyssey.
\newblock {\em IEEE Transactions on Neural Networks and Learning Systems},
  28(10):2222--2232, 2016.

\bibitem{resnet}
K. He, X. Zhang, S. Ren, and J. Sun.
\newblock Deep residual learning for image recognition.
\newblock In {\em Conference on Computer Vision and Pattern Recognition}, pages
  770--778, 2016.

\bibitem{LSTM}
S. Hochreiter and J. Schmidhuber.
\newblock Long short-term memory.
\newblock {\em Neural computation}, 9(8):1735--1780, 1997.

\bibitem{Lytro}
Lytro Inc.
\newblock Lytro illum.
\newblock \url{https://www.lytro.com}, 2015.

\bibitem{LipAAAI}
Y. Kumar, R. Jain, K. Salik, R. Shah, Y. Yin, and R. Zimmermann.
\newblock Lipper: Synthesizing thy speech using multi-view lipreading.
\newblock In {\em AAAI Conference on Artificial Intelligence}, pages
  10023--10024, 2019.

\bibitem{ACCV2}
D. Lee, J. Lee, and K. Kim.
\newblock Multi-view automatic lip-reading using neural network.
\newblock In {\em Asian Conference on Computer Vision}, pages 290--302, 2016.

\bibitem{LF}
M. Levoy and P. Hanrahan.
\newblock Light field rendering.
\newblock In {\em SIGGRAPH}, pages 31--42, 1996.

\bibitem{RNNSurvey}
Z. Lipton, J. Berkowitz, and C. Elkan.
\newblock A critical review of recurrent neural networks for sequence learning.
\newblock {\em arXiv:1506.00019}, 2015.

\bibitem{LSTMact2}
J. {Liu}, A. {Shahroudy}, D. {Xu}, A.~C. {Kot}, and G. {Wang}.
\newblock Skeleton-based action recognition using spatio-temporal {LSTM}
  network with trust gates.
\newblock {\em IEEE Transactions on Pattern Analysis and Machine Intelligence},
  40(12):3007--3021, 2018.

\bibitem{umap}
L. McInnes, J. Healy, and J. Melville.
\newblock {UMAP}: Uniform manifold approximation and projection for dimension
  reduction.
\newblock {\em arXiv:1802.03426}, 2018.

\bibitem{init}
N. Mohajerin and S. Waslander.
\newblock State initialization for recurrent neural network modeling of
  time-series data.
\newblock In {\em International Joint Conference on Neural Networks}, pages
  2330--2337, 2017.

\bibitem{BMVC}
S. Petridis, Y. Wang, Z. Li, and M. Pantic.
\newblock End-to-end multi-view lipreading.
\newblock {\em arXiv:1709.00443}, 2017.

\bibitem{LIP3}
S. Petridis, Y. Wang, Z. Li, and M. Pantic.
\newblock End-to-end multi-view lipreading.
\newblock In {\em The British Machine Vision Conference}, pages 1--14, 2017.

\bibitem{MVLSTM}
S. Rajagopalan, L. Morency, T. Baltrusaitis, and R. Goecke.
\newblock Extending long short-term memory for multi-view structured learning.
\newblock In {\em European Conference on Computer Vision}, pages 338--353,
  2016.

\bibitem{b17}
T. Rockt{\"a}schel, E. Grefenstette, K. Hermann, T. Ko{\v{c}}isk{\`y}, and P.
  Blunsom.
\newblock Reasoning about entailment with neural attention.
\newblock In {\em International Conference on Learning Representations}, pages
  1--9, 2016.

\bibitem{atlstm}
D. Sahrawat, Y. Kumar, S. Aggarwal, Y. Yin, R. Shah, and R. Zimmermann.
\newblock "notic my speech"--blending speech patterns with multimedia.
\newblock {\em arXiv:2006.08599}, 2020.

\bibitem{joint}
A. Sepas-Moghaddam, A. Etemad, F. Pereira, and P. Correia.
\newblock Long short-term memory with gate and state level fusion for light
  field-based face recognition.
\newblock {\em IEEE Transactions on Information Forensics and Security},
  16(1):1365--1379, 2020.

\bibitem{TIP}
A. Sepas-Moghaddam, A. Etemad, F. Pereira, and P. Correia.
\newblock Capsfield: Light field-based face and expression recognition in the
  wild using capsule routing.
\newblock {\em IEEE Transactions on Image Processing}, 30(1):2627--2642, 2021.

\bibitem{CSVT}
A. Sepas-Moghaddam, M. Haque, P. Correia, K. Nasrolahi, T. Moeslund, and F.
  Pereira.
\newblock A double-deep spatio-angular learning framework for light field based
  face recognition.
\newblock {\em IEEE Transactions on Circuits and Systems for Video Technology},
  30(12):4496--4512, 2020.

\bibitem{CRV}
J. Wang, X. Nie, Y. Wu, and S. Zhu.
\newblock Cross-view action modeling, learning and recognition.
\newblock In {\em Conference on Computer Vision and Pattern Recognition}, pages
  2649--2656, 2014.

\bibitem{dualLSTM}
J. Wang, M. Xue, R. Culhane, E. Diao, J. Ding, and V. Tarokh.
\newblock Speech emotion recognition with dual-sequence {LSTM} architecture.
\newblock In {\em Conference on Acoustics, Speech and Signal Processing}, pages
  6474--6478, 2020.

\bibitem{fuse8}
L. Wang, X. Zhao, and Y. Liu.
\newblock Skeleton feature fusion based on multi-stream {LSTM} for action
  recognition.
\newblock {\em IEEE Access}, 6:50788--50800, 2018.

\bibitem{ACCV}
M. Zimmermann, M.~Mehdipour Ghazi, H. Ekenel, and J. Thiran.
\newblock Visual speech recognition using {PCA} networks and {LSTMs} in a
  tandem {GMM-HMM} system.
\newblock In {\em Asian Conference on Computer Vision}, pages 264--276, 2016.

\end{thebibliography}
}

\end{document}